\documentclass[lettersize,journal]{IEEEtran}

\usepackage{amsmath,amsfonts}
\usepackage{algorithmic}
\usepackage{algorithm}
\usepackage{array}
\usepackage[caption=false,font=normalsize,labelfont=sf,textfont=sf]{subfig}
\usepackage{textcomp}
\usepackage{stfloats}
\usepackage{url}
\usepackage{verbatim}
\usepackage{graphicx}
\usepackage{cite}
\usepackage{booktabs}
\usepackage{multirow}
\usepackage{longtable}
\usepackage{adjustbox}
\usepackage{tabularx}
\usepackage{orcidlink}
\usepackage{xcolor}
\usepackage{capt-of}
\usepackage{hyperref}

\hypersetup{
    colorlinks=true,
    citecolor=blue,
    linkcolor=blue,
    urlcolor=blue
}

\begin{document}

\title{How Reliable Are Object Detection Models Beyond Clear Weather? A Deployment-Oriented Robustness Study for Autonomous Driving}

\author{
Munshaibur Rahman Mahin\,\orcidlink{0009-0007-4885-213X},
Isha Das\,\orcidlink{0009-0007-7812-4627},
Most.\ Sonia Islam\,\orcidlink{0009-0001-5930-6704},
Mohammad Sakib Mahmood\,\orcidlink{0000-0002-9198-9488},\\
M. F. Mridha\,\orcidlink{0000-0001-5738-1631}, \IEEEmembership{Senior Member, IEEE},
Md. Jakir Hossen\,\orcidlink{0000-0002-9978-7987}, \IEEEmembership{Senior Member, IEEE}
\thanks{Manuscript received MM DD, YYYY; revised MM DD, YYYY; accepted MM DD, YYYY. Date of publication MM DD, YYYY; current version MM DD, YYYY. This work was supported in part by XXXXXX. {\textit{(Corresponding authors: M.~F.~Mridha and Md.~Jakir Hossen.)}}.}

\thanks{The authors are Munshaibur Rahman Mahin (e-mail: skmahin661@gmail.com) department of computer science \& engineering, SEC, Shahjalal University of Science and Technology (SUST), Sylhet, Bangladesh. Isha Das (e-mail: ishadas2006@gmail.com) department of electrical engineering the University of Toronto, Toronto, ON, Canada. Most.\ Sonia Islam (e-mail: saniaislamsava@gmail.com) department of computer science \& engineering,  Bangladesh University of Business and Technology (BUBT), Dhaka, Bangladesh. Mohammad Sakib Mahmood (e-mail: mm974s@missouristate.edu) is with the department of computer science, Missouri State University, Springfield, MO, USA. M.~F.~Mridha (e-mail:  firoz.mridha@aiub.edu) is with the department of computer Science \& engineering, American International University-Bangladesh (AIUB), Dhaka, Bangladesh. Md.~Jakir Hossen (e-mail: jakir.hossen@mmu.edu.my) is with the Department of Robotics and Automation, Multimedia University, Melaka, Malaysia.}
}



\maketitle
\begin{abstract}
Autonomous driving perception systems must maintain reliable object detection performance under diverse environmental conditions rather than under clear daytime scenes alone. However, the cross-weather robustness of modern detection architectures remains insufficiently characterised under deployment-realistic source-only evaluation settings. This paper presents a controlled source-only cross-weather robustness study in which representative object detection architectures are trained exclusively on clear daytime driving scenes and evaluated without adaptation under unseen fog, rain, snow, and night environments. The evaluation includes lightweight one-stage, medium-scale one-stage, two-stage, and transformer-based detectors within a unified testing protocol. Detection robustness is quantified using performance Drop (\%) and a Weather Robustness Index (WRI), together with class-level sensitivity analysis for critical road participants including car, bus, truck, and person categories. Prediction confidence reliability under environmental domain shift is further analysed using Expected Calibration Error (ECE). Experimental results show consistent degradation across adverse environments, with performance dropping by up to 62.90\% under night-time conditions. Bus instances exhibit the highest sensitivity to environmental variability, whereas cars remain comparatively stable across weather transitions. Architecture-level comparison further indicates that YOLO-based detectors exhibit the lowest calibration mismatch, indicating stronger agreement between predicted detection confidence and empirical detection correctness, whereas DETR produces the lowest robustness score with a WRI of 0.6859, highlighting increased vulnerability under cross-weather domain shift. These findings provide deployment-oriented insights into architecture-level robustness behaviour for safety-critical intelligent vehicle perception systems.
\end{abstract}

\vspace{2cm}

\section{Introduction}

\IEEEPARstart{R}{eliable} visual perception is a fundamental requirement for autonomous driving systems operating in complex and continuously changing real-world environments \cite{10802028,9127855}. Among the core perception components, object detection plays a central role in identifying critical road participants such as cars, buses, trucks, and pedestrians, which directly influence scene understanding, trajectory planning, and safe driving decisions. Although autonomous driving platforms often integrate multiple sensing modalities such as LiDAR and radar \cite{10547458}, camera-based perception remains essential for recognising traffic participants and interpreting scene semantics required for safe navigation \cite{florea2022enhanced}.

Recent advances in deep learning–based object detection have substantially improved detection accuracy under favourable imaging conditions, particularly in clear daytime driving scenes, supporting the feasibility of deploying vision-based perception systems within intelligent vehicle platforms \cite{GUPTA2021100057}. Despite these improvements, the robustness of modern object detection models under environmental domain shift remains insufficiently characterised. Large-scale driving datasets \cite{11249750,seita2018bdd100k,kenk2020dawn} commonly used for detector training are dominated by clear daytime scenes, whereas adverse-weather conditions are comparatively under-represented. Consequently, models optimised under favourable environments may experience substantial performance degradation when deployed in previously unseen conditions such as fog, rain, snow, and night-time illumination changes, where reduced visibility and lighting variation introduce significant challenges for stable object localisation. Systematic evaluation under such environmental variability is therefore necessary for understanding detection reliability in safety-critical deployment scenarios.

Existing studies on adverse-weather perception have largely focused on improving detection performance through domain adaptation strategies, synthetic weather augmentation, or specialised architectural modifications \cite{li2023domain}. Although these approaches can improve accuracy, they modify the training process by incorporating target-domain information or artificially generated environmental conditions and therefore do not directly reveal how well detection models inherently generalise beyond the conditions observed during training \cite{10918696}. A controlled source-only evaluation protocol is therefore necessary to isolate environmental domain-shift effects and enable deployment-oriented benchmarking of intrinsic robustness under realistic operating conditions.

Motivated by this need, this study investigates how representative object detection models perform when trained exclusively on clear daytime driving scenes and evaluated under unseen adverse-weather environments including fog, rain, snow, and night conditions. By excluding adverse-weather samples from training, the experimental design isolates performance degradation caused solely by environmental domain shift and enables controlled benchmarking of intrinsic cross-weather robustness across detector architectures within a unified source-only evaluation framework. In addition to detection accuracy and robustness degradation, this study further examines prediction confidence reliability under environmental variation using calibration analysis, which is essential for safe deployment of perception systems operating in uncertain real-world conditions.

Within this controlled experimental protocol, the study addresses the following research questions:

\begin{itemize}

\item \text{RQ1:} How does object detection performance degrade when models trained exclusively on clear daytime driving scenes are evaluated under unseen adverse-weather conditions such as fog, rain, snow, and night environments?

\item \text{RQ2:} Which adverse-weather conditions produce the greatest degradation in detection performance under cross-weather domain shift?

\item \text{RQ3:} Which object categories, including car, bus, truck, and person, are most vulnerable to weather-induced performance degradation?

\item \text{RQ4:} How do different detection architectures compare in terms of robustness retention under cross-weather domain shift when assessed using Drop (\%) and the Weather Robustness Index (WRI)?

\end{itemize}

By addressing these questions, this study provides a structured evaluation of cross-weather detection reliability using a controlled source-only training protocol. Rather than proposing a new detection architecture or training strategy, the work presents a deployment-oriented robustness benchmarking study that analyses architecture-level robustness, object-category sensitivity, and confidence reliability under unseen adverse-weather conditions relevant to safety-critical autonomous driving perception systems.

The main contributions of this study are summarised as follows:

\begin{itemize}

\item A controlled source-only cross-weather evaluation protocol based on training with clear daytime scenes from BDD100K \cite{bdd100k_dataset} and testing on unseen adverse-weather conditions from ACDC \cite{acdc_dataset_website}, enabling direct measurement of environmental domain-shift effects without domain adaptation or synthetic augmentation.

\item A systematic robustness analysis using deployment-oriented indicators, including performance Drop (\%) and the Weather Robustness Index (WRI), to quantify localisation stability under adverse-weather conditions.

\item A confidence calibration evaluation using Expected Calibration Error (ECE) to assess prediction reliability across environmental domain-shift scenarios.

\item A comprehensive sensitivity analysis across adverse-weather conditions and object categories (car, bus, truck, person), together with an architecture-level robustness comparison across representative one-stage, two-stage, and transformer-based detectors.
\end{itemize}

The remainder of this paper is organised as follows. Section~\ref{sec:literature_review} reviews the relevant literature. Section~\ref{sec:methodology} describes the datasets, experimental protocol, and robustness metrics. Section~\ref{sec:results} presents the experimental results and analysis. Section~\ref{sec:conclusion} summarises the findings and outlines future work.

\section{Literature Review}
\label{sec:literature_review}
The development of autonomous driving perception systems has progressed from early modular vision-based frameworks to modern deep learning detection architectures capable of real-time scene understanding \cite{Kalakunnath_Varghese_Anuradha_Girish_Sasidharan_2025}.

Bücher et al. \cite{buecher2001intelligent} introduced an early hierarchical perception pipeline integrating lane and obstacle detection with world-coordinate mapping for behaviour planning. With the emergence of convolutional detection architectures, Ren et al.  \cite{ren2015faster} proposed Faster R-CNN, enabling efficient region proposal generation for accurate two-stage detection. In parallel, Redmon et al. \cite{redmon2016yolo} reformulated detection as a unified regression task through the YOLO architecture, enabling real-time perception with global contextual reasoning. Later improvements further enhanced efficiency–accuracy trade-offs; Bochkovskiy et al. \cite{bochkovskiy2020yolov4} introduced YOLOv4 using optimised training strategies for deployment-oriented performance. Complementing convolutional detectors, Carion et al. \cite{carion2020detr} proposed DETR, a transformer-based architecture enabling global relational reasoning without handcrafted proposal mechanisms. Despite these advances, the robustness of modern detection architectures under adverse-weather deployment conditions remains insufficiently characterised.

Environmental variability continues to challenge perception reliability in autonomous driving. Although large-scale datasets improve scene diversity, adverse-weather conditions remain comparatively under-represented. Yu et al. \cite{yu2020bdd100k} showed through the BDD100K dataset that detection performance varies across environmental contexts, particularly between daytime and nighttime scenes. Similarly, Sakaridis et al. (b) \cite{sakaridis2021acdc} introduced the ACDC dataset to highlight visually indiscernible regions under fog, rain, snow, and night environments that complicate camera-based perception. Gupta et al.  \cite{gupta2024robust} further demonstrated that atmospheric disturbances such as rain streaks and snowfall degrade feature consistency and localisation reliability. These findings emphasise the need for systematic evaluation of detection robustness under deployment-relevant environmental variability.

To overcome these issues, the past research has mostly.
concerned about enhanced adverse-weather performance by means of.
domain adaptation and weather-aware augmentation approaches.
Zhu et al. \cite{zhu2017cyclegan} proposed CycleGAN to unpaired image-toimage translation between visual domains and is synthetic
change in appearance in the course of training. Chen et al. \cite{chen2018domain}
introduced domain adaptive Faster R-CNN, to lessen performance decline due to distribution mismatch between source domain and target domain. Likewise, Sakaridis et al. (a) \cite{sakaridis2018semantic}
has shown that synthetic fog simulation can enhance segmentation
performance of in poor visibility.
Recently, Hoyer et al. \cite{hoyer2022daformer} proposed DAFormer, which demonstrates.
the context-aware feature fusion based on transformers is enhanced adaptation to non-homogeneous environments. Nevertheless, all of these methods manipulate the training distribution, and thus do not directly indicate how detection models naturally generalise during source-only training when exposed to unknown weather conditions.

In addition to adaptation-based strategies, a number of
research on robustness benchmarking with distribution shift has been examined. ImageNetC benchmark to measure algorithmically sensitivity of models was presented by Hendrycks et al. \cite{hendrycks2019benchmarking}.
generated corruptions. Geirhos et al. \cite{geirhos2019imagenet} also demonstrated that
texture bias is the limit to resilience to distribution shift, and
shape-biased training enhances generalisation. Focusing on
detection tasks, Michaelis et al. \cite{michaelis2019benchmarking} suggested the Robust.
Detection Benchmark and recorded high performance.
degradation on invisible corrupted realms. Similarly, Bijelic
et al. \cite{bijelic2020seeing} showed that degradation was measurable when models.
trained on clean data are subject to poor visibility like heavy snowfall and fog.
Despite these findings, structured cross-weather robustness evaluation under controlled source-only training conditions remains insufficiently explored for modern autonomous driving object detection architectures, particularly with deployment-oriented robustness indicators that explicitly quantify performance degradation under environmental domain shift.

\begin{figure}[ht]
\centering
\includegraphics[width=\columnwidth]{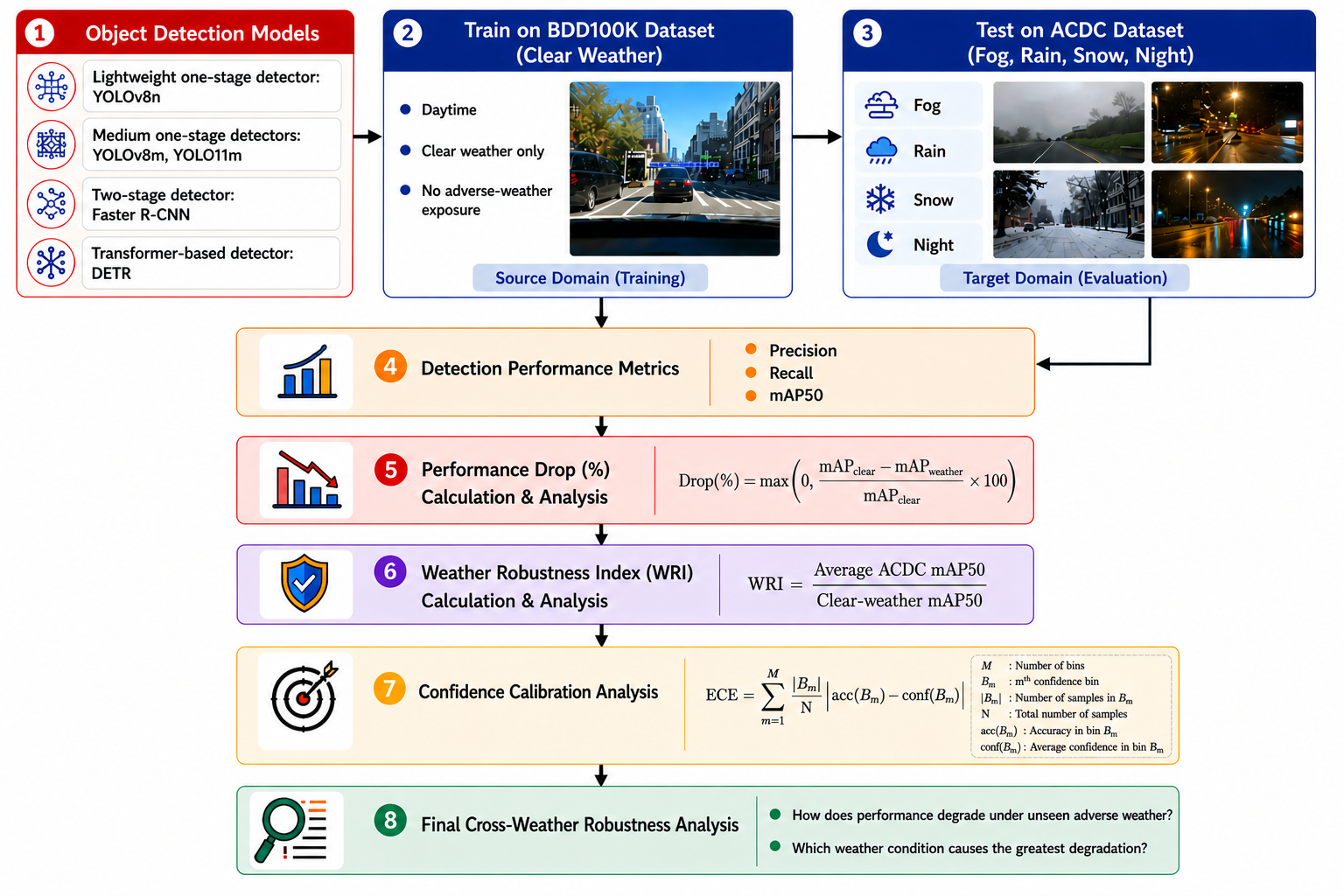}
\caption{Source-only cross-weather robustness evaluation workflow }
\label{fig:workflow}
\end{figure}

\section{Methodology}
\label{sec:methodology}
This section describes the experimental configuration used
to evaluate cross-weather robustness under a controlled source-
only training protocol. Models were trained on clear day-
time scenes and evaluated across multiple adverse-weather
conditions using standard detection metrics and deployment-
oriented robustness indicators. The following subsections
present the datasets, training settings, and evaluation proce-
dures adopted in this study, while Fig.~\ref{fig:workflow}
illustrates the overall workflow of the proposed cross-weather
robustness evaluation framework.

\subsection{Dataset Description}
This study evaluates object detection robustness under environmental domain shift in autonomous driving perception using the BDD100K dataset \cite{bdd100k_dataset} as the source domain and the ACDC dataset \cite{acdc_dataset_website} as the target domain. Detection models are trained exclusively on clear daytime driving scenes and evaluated under previously unseen adverse-weather environments to measure robustness under deployment-oriented environmental variability.

The BDD100K dataset was originally collected as vehicle-mounted driving video sequences with object detection annotations provided at the frame level. In this study, annotated frames captured under clear daytime conditions are used directly for training. The four categories of traffic-relevant objects are stored: car, bus, truck, person. The dataset is trained, validated, and tested according to its predefined partitions to make the model selection reproducible and consistent.
The distribution of the classes in the BDD dataset is inherently unbalanced, and car occurrences are more common compared to bus, trucks, and pedestrians. Since the objective of this study is to evaluate cross-weather robustness under realistic deployment conditions rather than optimise class-balanced detection performance, the original dataset distribution was preserved to reflect practical autonomous driving scenarios.

Cross-weather robustness evaluation is performed using the ACDC dataset, which provides annotated driving scenes captured under fog, rain, snow, and night conditions. These subsets represent distinct visibility degradation environments and are used exclusively as unseen target domains during evaluation. The same four object categories are preserved across both datasets to maintain consistent category alignment between training and testing domains.

\begin{figure*}[ht]
\centering
\includegraphics[width=\textwidth]{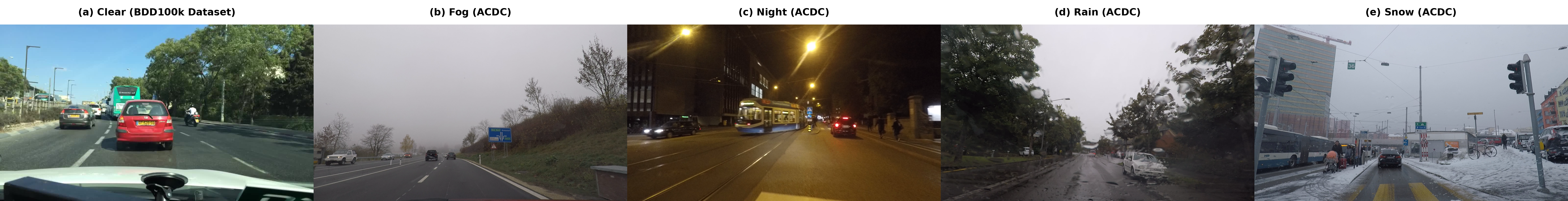}
\caption{Source and target dataset examples for cross-weather evaluation: BDD100K clear scenes for training and ACDC adverse-weather scenes (fog, night, rain, snow) for evaluation.}
\label{fig:dataset_weather_examples}
\end{figure*}

A summary of dataset composition across source and target domains is provided in Table~\ref{tab:dataset_statistics} and to illustrate the environmental domain-shift evaluation protocol used in this study, representative driving scenes from the source and target datasets are shown in Fig.~\ref{fig:dataset_weather_examples}. 

\begin{table*}[ht]
\centering
\caption{Dataset composition of BDD100K (source) and ACDC (target) across splits and weather conditions for the four object classes.}
\label{tab:dataset_statistics}
\begin{tabular}{llllllll}
\toprule
Dataset & Domain & Condition & Images & Car & Bus & Truck & Person \\
\midrule

\multirow{3}{*}{BDD100K} 
& \multirow{3}{*}{Source} 
& Clear (Train)      & 12,424 & 139,549 & 2,732 & 7,424 & 16,784 \\
& 
& Clear (Validation) & 1,757  & 19,639  & 380   & 1,022 & 2,451  \\
& 
& Clear (Test)       & 3,563  & 39,711  & 754   & 2,205 & 4,242  \\

\multirow{4}{*}{ACDC} 
& \multirow{4}{*}{Target} 
& Fog        & 490    & 2,134   & 23    & 401   & 238    \\
& 
& Rain       & 1,661  & 7,197   & 200   & 553   & 1,826  \\
& 
& Snow       & 496    & 2,371   & 68    & 173   & 778    \\
& 
& Night      & 487    & 2,000   & 57    & 22    & 829    \\

\bottomrule
\end{tabular}
\end{table*}
\subsection{Experimental Protocol}

To evaluate cross-weather domain generalisation under controlled source-only conditions, five object detection architectures representing three major detector paradigms were selected: YOLOv8n as a lightweight one-stage detector; YOLOv8m and YOLO11m as medium-scale one-stage detectors; Faster R-CNN as a two-stage detector; and DETR as a transformer-based detection architecture. These models enable robustness behaviour to be examined across different localisation strategies within a unified evaluation framework.

All models were trained exclusively using clear daytime images from the source domain and evaluated without adaptation on adverse-weather environments. To ensure fair comparison across detector families, a consistent training budget of 100 epochs and identical input resolution (768 pixels) were adopted where applicable. AdamW optimisation with an initial learning rate of 0.001 was used across all architectures to maintain comparable optimisation conditions.

For Faster R-CNN and DETR, a batch size of 8 was used due to their higher computational complexity and memory requirements when trained at 768-pixel resolution. These settings ensured stable training under the available hardware configuration (NVIDIA RTX A4000 GPU) while preserving consistency with the unified training protocol applied across detector architectures.

To ensure reproducibility across all experiments, deterministic training conditions were enforced using a fixed random seed of 42 for dataset handling, model initialisation, and optimisation procedures. A summary of training configurations across detector architectures is provided in Table~\ref{tab:training_configurations}.

\begin{table*}[ht]
\centering
\caption{Training configuration across the detection architectures used in this study}
\label{tab:training_configurations}

\begin{tabular}{lllllll}
\toprule
Architecture Type & Detector & Epochs & Image Size & Batch Size & Optimiser & Learning Rate \\
\midrule

Lightweight one-stage
& YOLOv8n & 100 & 768 & 20 & AdamW & 0.001 \\

\multirow{2}{*}{Medium one-stage}
& YOLOv8m & 100 & 768 & 20 & AdamW & 0.001 \\
& YOLO11m & 100 & 768 & 20 & AdamW & 0.001 \\

Two-stage
& Faster R-CNN & 100 & 768 & 8 & AdamW & 0.001 \\

Transformer-based
& DETR & 100 & Default preprocessing & 8 & AdamW & 0.001 \\

\bottomrule
\end{tabular}

\end{table*}

\subsection{Evaluation Metrics and Robustness Indicators}
\label{sec:eval_metrics}
To evaluate cross-weather robustness under adverse driving conditions, detection performance was assessed using standard object detection metrics together with deployment-oriented robustness indicators and confidence calibration analysis. These measures support interpretation of detection accuracy, weather-induced degradation, and prediction reliability under environmental domain shift.

Detection accuracy was evaluated using mean Average Precision at an intersection-over-union threshold of 0.50 (mAP50), together with precision and recall to characterise localisation quality, detection completeness, and missed-object sensitivity. All metrics were reported both class-wise (car, bus, truck, person) and as macro-averaged values to enable balanced comparison across object categories.

\subsubsection{Performance Degradation Metric (Drop (\%))}

To quantify robustness under unseen adverse-weather conditions beyond in-domain accuracy evaluation, performance degradation was measured using the Drop (\%) metric. This metric represents the relative reduction in detection performance between the clear-weather source domain and adverse-weather target domains and was computed as

\begin{equation}
\mathrm{Drop}(\%) =
\max \left(0,\;
\frac{\mathrm{mAP}_{\text{clear}} -
\mathrm{mAP}_{\text{weather}}}
{\mathrm{mAP}_{\text{clear}}}
\times 100
\right)
\label{eq:drop_metric}
\end{equation}

The use of the $\max(0,\mathrm{Drop})$ formulation ensures that only performance reductions are interpreted as degradation, while occasional improvements under specific weather conditions are not treated as negative robustness effects. This provides a consistent measure of environmental domain-shift impact across detection architectures.

\subsubsection{Weather Robustness Index (WRI)}

To support architecture-level comparison of cross-weather robustness, a Weather Robustness Index (WRI) was used as a deployment-oriented robustness indicator defined as

\begin{equation}
\mathrm{WRI} =
\frac{\text{Average ACDC mAP50}}
{\text{Clear-weather mAP50}}
\label{eq:wri_metric}
\end{equation}

Unlike individual Drop (\%) values that quantify degradation separately for each weather condition, WRI summarises the overall proportion of detection performance retained across adverse-weather environments relative to the clear-weather baseline. For interpretability within this evaluation framework, WRI values were categorised as shown in Table~\ref{tab:wri_scale}.

\begin{table}[ht]
\centering
\caption{Interpretation guideline for Weather Robustness Index (WRI) values}
\label{tab:wri_scale}
\begin{tabular}{cc}
\hline
\text{WRI Range} & \text{Robustness Interpretation} \\
\hline
$\geq 0.90$ & Highly robust performance \\
0.80 - 0.89 & Moderately robust performance \\
0.70 - 0.79 & Moderately weak \\
$< 0.70$ & Severe performance degradation (weak) \\
\hline
\end{tabular}
\end{table}

Together, these evaluation measures enable detector behaviour to be analysed from complementary perspectives including absolute detection accuracy, weather-specific degradation magnitude, and overall robustness consistency across adverse-weather environments.

\subsubsection{Confidence Calibration Analysis Using Expected Calibration Error (ECE)}

In addition to detection accuracy and robustness indicators, prediction confidence reliability was evaluated using Expected Calibration Error (ECE). Calibration analysis measures the degree of agreement between predicted detection confidence scores and empirical detection correctness, providing insight into how reliably a detector estimates its own prediction certainty under environmental domain shift conditions.

In safety-critical autonomous driving scenarios, reliable confidence estimation is essential because perception systems must not only detect objects accurately but also quantify prediction uncertainty consistently across changing visibility conditions. Models that produce overconfident predictions under adverse weather may introduce increased deployment risk even when detection accuracy remains acceptable.

ECE was computed by partitioning prediction confidence scores into $M$ equal-width bins and comparing the average confidence and empirical accuracy within each bin. A predicted detection was considered correct when its predicted class matched the corresponding ground-truth label and its intersection-over-union (IoU) with the ground-truth bounding box exceeded 0.50. The ECE metric was calculated as

\begin{equation}
\mathrm{ECE} =
\sum_{m=1}^{M}
\frac{|B_m|}{N}
\left|
\mathrm{acc}(B_m) -
\mathrm{conf}(B_m)
\right|
\label{eq:ece_metric}
\end{equation}

where $B_m$ denotes the set of detections whose confidence scores fall into bin $m$, $|B_m|$ represents the number of detections in that bin, $N$ is the total number of detections, $\mathrm{acc}(B_m)$ denotes the empirical detection accuracy within the bin, and $\mathrm{conf}(B_m)$ denotes the average predicted confidence within the bin.

Lower ECE values indicate stronger agreement between prediction confidence and actual detection correctness, reflecting improved reliability of confidence estimation under cross-weather evaluation conditions.
\section{Results and Discussion}
\label{sec:results}
This section presents the experimental evaluation of cross-weather detection robustness when models trained on clear daytime scenes are deployed under unseen adverse-weather environments. Performance is analysed through in-domain baseline accuracy, cross-weather localisation behaviour, architecture-level degradation using Drop (\%), robustness ranking using the Weather Robustness Index (WRI), object-category sensitivity analysis, and confidence reliability assessment using Expected Calibration Error (ECE). Together, these evaluations provide a structured interpretation of how environmental domain shift affects detection reliability across detector architectures and traffic participants.
\begin{table}[t]
\centering
\caption{Object-wise in-domain detection performance on clear-weather BDD100K test set}
\label{tab:baseline_classwise_results}

\begin{tabular}{llccc}
\toprule
Model & Object/Class & Precision & Recall & mAP50 \\
\midrule

\multirow{4}{*}{YOLOv8n}
& car    & 0.8157 & 0.7040 & 0.7760 \\
& bus    & 0.6655 & 0.4881 & 0.5370 \\
& truck  & 0.6262 & 0.5038 & 0.5640 \\
& person & 0.6936 & 0.4455 & 0.5317 \\

\midrule

\multirow{4}{*}{YOLOv8m}
& car    & 0.8466 & 0.7553 & 0.8233 \\
& bus    & 0.6967 & 0.5454 & 0.6038 \\
& truck  & 0.6505 & 0.5828 & 0.6255 \\
& person & 0.7398 & 0.5443 & 0.6266 \\

\midrule

\multirow{4}{*}{YOLO11m}
& car    & 0.8532 & 0.7586 & 0.8314 \\
& bus    & 0.6727 & 0.5650 & 0.6227 \\
& truck  & 0.6738 & 0.5896 & 0.6343 \\
& person & 0.7528 & 0.5523 & 0.6386 \\

\midrule

\multirow{4}{*}{Faster R-CNN}
& car    & 0.4835 & 0.8063 & 0.7560 \\
& bus    & 0.4232 & 0.5849 & 0.5083 \\
& truck  & 0.4326 & 0.5615 & 0.4787 \\
& person & 0.3489 & 0.5922 & 0.4946 \\

\midrule

\multirow{4}{*}{DETR}
& car    & 0.4046 & 0.7617 & 0.3889 \\
& bus    & 0.5027 & 0.3714 & 0.2593 \\
& truck  & 0.3120 & 0.4866 & 0.2826 \\
& person & 0.3381 & 0.5498 & 0.1870 \\

\bottomrule
\end{tabular}

\end{table}

\begin{table*}[!t]
\centering
\makebox[\textwidth][c]{%

\begin{minipage}[t]{0.49\textwidth}
\centering
\caption{Object-wise cross-weather detection performance of YOLO-based detectors on ACDC adverse-weather conditions}
\label{tab:yolo_cross_weather_detection}

\resizebox{\linewidth}{!}{
\begin{tabular}{l l p{1.1cm} c c c}
\toprule
Model & Weather & Object/Class & Precision & Recall & mAP50 \\
\midrule

\multirow{16}{*}{YOLOv8n}
& Fog & car & 0.7822 & 0.7184 & 0.7743 \\
&  & bus & 0.2746 & 0.7391 & 0.6561 \\
&  & truck & 0.7372 & 0.4165 & 0.5372 \\
&  & person & 0.6151 & 0.3950 & 0.4626 \\
\cmidrule(lr){2-6}

& Rain & car & 0.7081 & 0.6779 & 0.7211 \\
&  & bus & 0.3577 & 0.2850 & 0.2002 \\
&  & truck & 0.5315 & 0.4014 & 0.3916 \\
&  & person & 0.6719 & 0.3364 & 0.3877 \\
\cmidrule(lr){2-6}

& Snow & car & 0.7570 & 0.7221 & 0.7883 \\
&  & bus & 0.4718 & 0.2941 & 0.2324 \\
&  & truck & 0.4467 & 0.3815 & 0.3373 \\
&  & person & 0.7658 & 0.3946 & 0.4841 \\
\cmidrule(lr){2-6}

& Night & car & 0.6159 & 0.5204 & 0.5620 \\
&  & bus & 0.2900 & 0.0877 & 0.0652 \\
&  & truck & 0.1474 & 0.2273 & 0.0855 \\
&  & person & 0.7327 & 0.2509 & 0.3108 \\

\midrule

\multirow{16}{*}{YOLOv8m}
& Fog & car & 0.8301 & 0.7532 & 0.8170 \\
&  & bus & 0.3638 & 0.7391 & 0.6505 \\
&  & truck & 0.8045 & 0.4838 & 0.6461 \\
&  & person & 0.7046 & 0.4580 & 0.5614 \\
\cmidrule(lr){2-6}

& Rain & car & 0.7596 & 0.7249 & 0.7726 \\
&  & bus & 0.3793 & 0.3250 & 0.2876 \\
&  & truck & 0.5625 & 0.4626 & 0.4842 \\
&  & person & 0.7245 & 0.4135 & 0.4764 \\
\cmidrule(lr){2-6}

& Snow & car & 0.7851 & 0.7735 & 0.8254 \\
&  & bus & 0.3183 & 0.3235 & 0.2699 \\
&  & truck & 0.4225 & 0.3931 & 0.4004 \\
&  & person & 0.7394 & 0.4814 & 0.5588 \\
\cmidrule(lr){2-6}

& Night & car & 0.7590 & 0.5540 & 0.6387 \\
&  & bus & 0.3662 & 0.1228 & 0.1412 \\
&  & truck & 0.1315 & 0.2273 & 0.0796 \\
&  & person & 0.7781 & 0.3172 & 0.4060 \\

\midrule

\multirow{16}{*}{YOLO11m}
& Fog & car & 0.8413 & 0.7605 & 0.8261 \\
&  & bus & 0.3873 & 0.8245 & 0.6932 \\
&  & truck & 0.8260 & 0.4913 & 0.6585 \\
&  & person & 0.7239 & 0.4538 & 0.5733 \\
\cmidrule(lr){2-6}

& Rain & car & 0.7618 & 0.7403 & 0.7835 \\
&  & bus & 0.3856 & 0.3860 & 0.2989 \\
&  & truck & 0.5871 & 0.4955 & 0.4990 \\
&  & person & 0.7190 & 0.4211 & 0.4880 \\
\cmidrule(lr){2-6}

& Snow & car & 0.8195 & 0.7731 & 0.8352 \\
&  & bus & 0.3392 & 0.3851 & 0.2600 \\
&  & truck & 0.4390 & 0.4277 & 0.3924 \\
&  & person & 0.8106 & 0.4731 & 0.5751 \\
\cmidrule(lr){2-6}

& Night & car & 0.6875 & 0.6020 & 0.6517 \\
&  & bus & 0.3199 & 0.1652 & 0.1392 \\
&  & truck & 0.1262 & 0.2273 & 0.0573 \\
&  & person & 0.7392 & 0.3624 & 0.4178 \\

\bottomrule
\end{tabular}
}
\end{minipage}

\hspace{0.02\textwidth}

\begin{minipage}[t]{0.49\textwidth}
\centering
\caption{Object-wise cross-weather detection performance Faster R-CNN and DETR detectors on ACDC adverse-weather conditions.}
\label{tab:cross_weather_detection_cont}

\resizebox{\linewidth}{!}{
\begin{tabular}{l l p{1.1cm} c c c}
\toprule
Model & Weather & Object/Class & Precision & Recall & mAP50 \\
\midrule

\multirow{16}{*}{Faster R-CNN}
& Fog & car & 0.3519 & 0.8276 & 0.7604 \\
&  & bus & 0.2125 & 0.7391 & 0.5817 \\
&  & truck & 0.5340 & 0.6658 & 0.6088 \\
&  & person & 0.3273 & 0.5294 & 0.4484 \\
\cmidrule(lr){2-6}

& Rain & car & 0.2619 & 0.7930 & 0.6988 \\
&  & bus & 0.2500 & 0.3950 & 0.2483 \\
&  & truck & 0.3176 & 0.5967 & 0.4697 \\
&  & person & 0.2829 & 0.4819 & 0.3924 \\
\cmidrule(lr){2-6}

& Snow & car & 0.2843 & 0.8397 & 0.7563 \\
&  & bus & 0.2180 & 0.4265 & 0.2797 \\
&  & truck & 0.2406 & 0.4798 & 0.3652 \\
&  & person & 0.3170 & 0.5566 & 0.4772 \\
\cmidrule(lr){2-6}

& Night & car & 0.1803 & 0.6850 & 0.5655 \\
&  & bus & 0.2059 & 0.2456 & 0.1017 \\
&  & truck & 0.0709 & 0.4091 & 0.1309 \\
&  & person & 0.3139 & 0.3896 & 0.3119 \\

\midrule

\multirow{16}{*}{DETR}
& Fog & car & 0.3354 & 0.7478 & 0.3461 \\
&  & bus & 0.2063 & 0.5652 & 0.3230 \\
&  & truck & 0.3981 & 0.4150 & 0.2500 \\
&  & person & 0.3769 & 0.5254 & 0.1657 \\
\cmidrule(lr){2-6}

& Rain & car & 0.2687 & 0.7268 & 0.3324 \\
&  & bus & 0.2314 & 0.2650 & 0.0818 \\
&  & truck & 0.2321 & 0.3986 & 0.1720 \\
&  & person & 0.3258 & 0.4357 & 0.1360 \\
\cmidrule(lr){2-6}

& Snow & car & 0.3235 & 0.7784 & 0.3870 \\
&  & bus & 0.1782 & 0.2647 & 0.0891 \\
&  & truck & 0.1949 & 0.3988 & 0.1841 \\
&  & person & 0.3538 & 0.5293 & 0.1850 \\
\cmidrule(lr){2-6}

& Night & car & 0.1812 & 0.6097 & 0.2550 \\
&  & bus & 0.1556 & 0.1228 & 0.0352 \\
&  & truck & 0.0580 & 0.3636 & 0.0293 \\
&  & person & 0.3461 & 0.3097 & 0.0951 \\

\bottomrule
\end{tabular}
}

\vspace{10pt}

\caption{Weather-wise macro mAP50 across ACDC adverse-weather conditions}
\label{tab:macro_cross_weather}

\label{tab:macro_cross_weather}

\setlength{\tabcolsep}{3pt}
\footnotesize

\resizebox{\linewidth}{!}{
\begin{tabular}{lcccccc}
\toprule
Model & Clear & Fog & Rain & Snow & Night & Avg ACDC \\
\midrule
YOLOv8n      & 0.6022 & 0.6076 & 0.4252 & 0.4605 & 0.2559 & 0.4373 \\
YOLOv8m      & 0.6698 & 0.6688 & 0.5052 & 0.5136 & 0.3164 & 0.5010 \\
YOLO11m      & 0.6818 & 0.6878 & 0.5174 & 0.5157 & 0.3165 & 0.5094 \\
Faster R-CNN & 0.5594 & 0.5998 & 0.4523 & 0.4696 & 0.2775 & 0.4498 \\
DETR         & 0.2795 & 0.2712 & 0.1806 & 0.2113 & 0.1037 & 0.1917 \\
\bottomrule
\end{tabular}
}

\end{minipage}
}
\end{table*}

\subsection{In-domain detection performance on clear-weather scenes}

Detection performance under clear-weather conditions provides the source-domain baseline for subsequent cross-weather robustness analysis. Table~\ref{tab:baseline_classwise_results} summarises object-wise detection results across representative one-stage, two-stage, and transformer-based architectures evaluated on the clear-weather test split of the BDD100K dataset \cite{bdd100k_dataset}.

Across all detectors, vehicle categories achieved higher localisation accuracy than pedestrian instances under favourable visibility conditions. Medium-scale one-stage detectors produced the strongest baseline performance, with YOLO11m achieving the highest mAP50 for the car class (0.8314) and consistently improved results across the remaining categories. Faster R-CNN showed comparatively higher recall but lower precision across several classes, while DETR demonstrated reduced localisation accuracy overall, particularly for the person category.

\subsection{Cross-weather robustness evaluation on ACDC adverse-weather environments}

\begin{figure}[ht]
\centering
\includegraphics[width=\columnwidth]{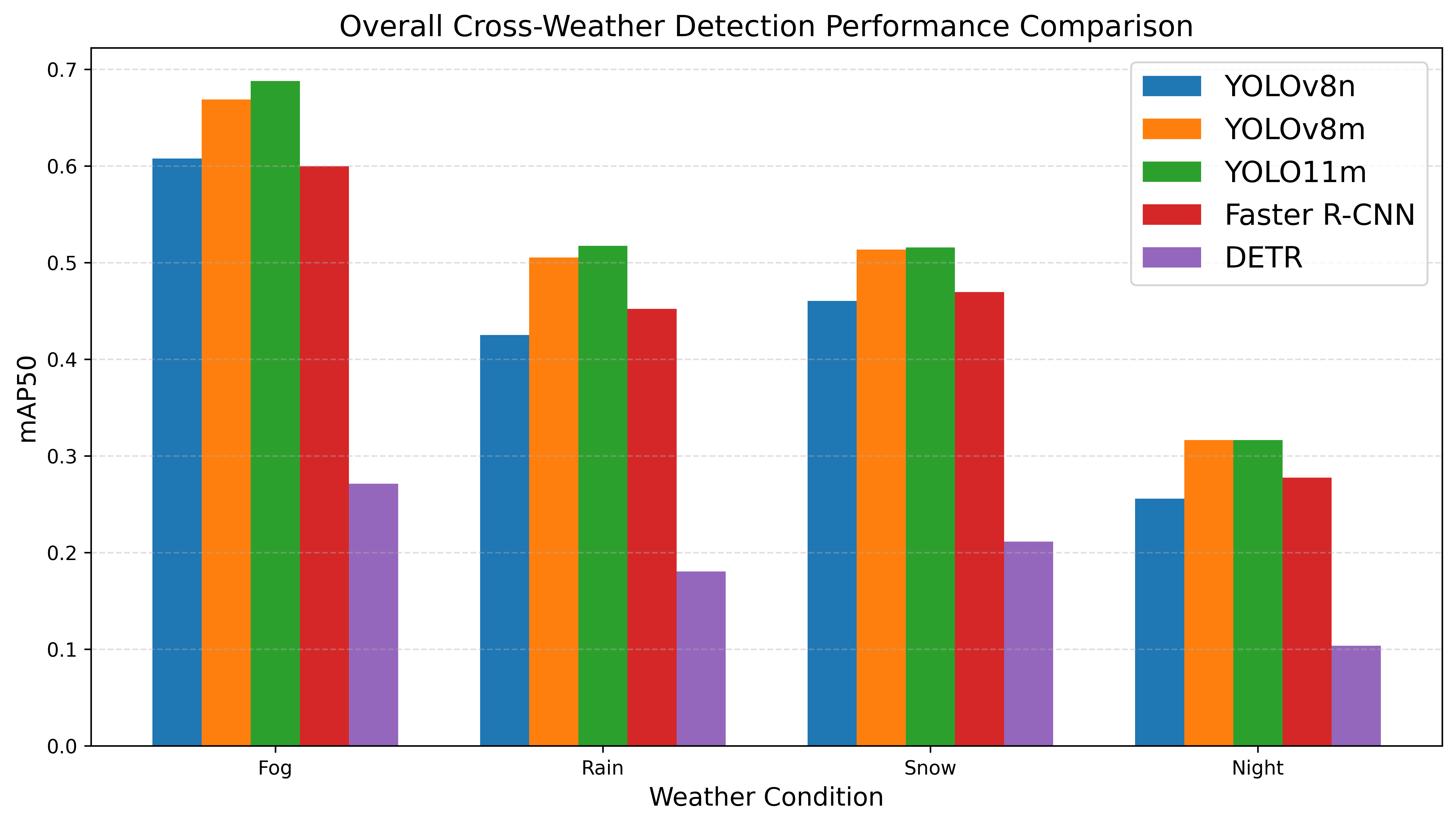}
\caption{Cross-weather detection performance (macro mAP50) under adverse conditions, worst at night.}
\label{fig:overall_cross_weather_performance}
\end{figure}

Cross-weather detection performance on ACDC dataset's \cite{acdc_dataset_website} adverse-weather environments is summarised in Tables~\ref{tab:yolo_cross_weather_detection} and \ref{tab:cross_weather_detection_cont}, with overall weather-wise performance trends illustrated in Fig.~\ref{fig:overall_cross_weather_performance}. The results reveal structured degradation patterns across environmental conditions when models trained exclusively on clear daytime BDD100K scenes are evaluated under fog, rain, snow, and night scenarios.

Across all evaluated conditions, the car category remained the most stable object class, with localisation performance largely preserved under fog and snow environments (mAP50 between 0.7743 and 0.8352) and remaining comparatively reliable under night conditions (0.5620–0.6517). This behaviour indicates that larger road participants retain discriminative structural cues despite atmospheric degradation.
In contrast, bus detection exhibited strong sensitivity to environmental variation, with localisation performance decreasing from 0.6561 under fog conditions to 0.2002 under rain and further to 0.0652 under night environments. A similar degradation pattern was observed across weather transitions, highlighting increased dependence of bus detection reliability on visibility and illumination conditions.

The truck category showed moderate robustness under fog environments (0.5372–0.6585), but experienced substantial degradation under night conditions, where localisation performance decreased to 0.0573. This behaviour reflects increased sensitivity of medium-scale vehicle structures to illumination variation and contrast reduction.
Detection of pedestrians remained comparatively stable under fog and snow conditions (0.4626–0.5751), but decreased under rain and night environments (0.3108–0.4178), indicating higher vulnerability of smaller object categories to reduced visibility and background complexity.
From an environmental perspective, Fig.~\ref{fig:overall_cross_weather_performance} shows that fog and snow conditions preserved comparatively higher detection stability, whereas rain introduced moderate degradation. Night environments produced the largest performance reductions across bus and truck categories, while pedestrian localisation became increasingly unstable under reduced visibility conditions. In contrast, car detection remained comparatively stable across most adverse-weather scenarios. These observations indicate that cross-weather performance degradation is influenced primarily by object scale, structural characteristics, and illumination variation rather than uniformly affecting all traffic participants.

To provide a compact summary of overall localisation behaviour across environmental conditions, the corresponding weather-wise macro mAP50 values are reported in Table~\ref{tab:macro_cross_weather}.

\subsection{Architecture-level degradation analysis using Drop (\%)}

\begin{figure}[ht]
\centering
\includegraphics[width=\columnwidth]{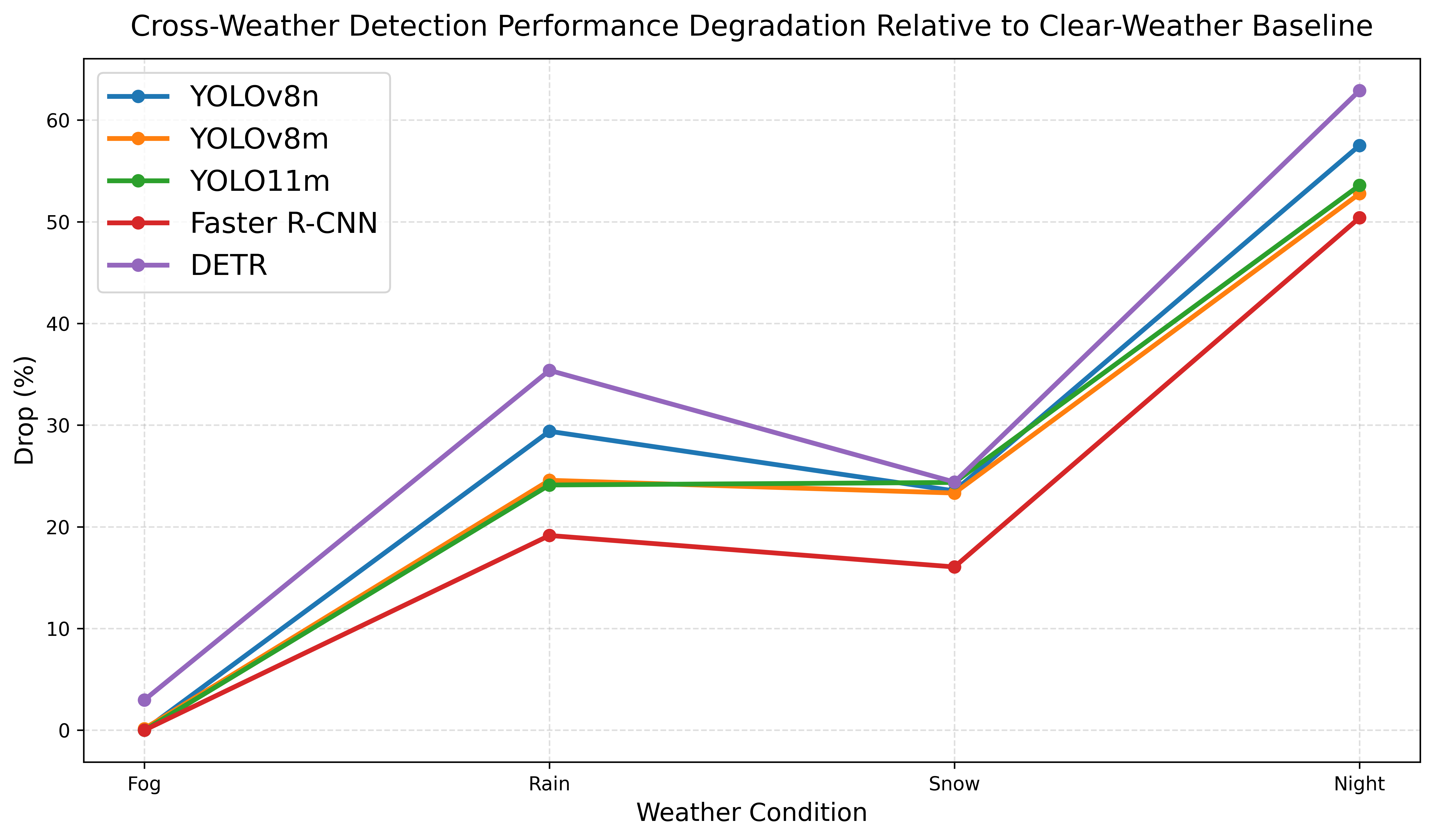}
\caption{Performance degradation relative to the clear-weather baseline (Drop, \%).}
\label{fig:drop_analysis_macro}
\end{figure}

To quantify robustness under environmental domain shift, performance degradation relative to the clear-weather baseline was measured using the Drop (\%) metric defined in
Eq.~(\ref{eq:drop_metric}) and reported in Table~\ref{tab:map50_drop_results}, with the corresponding trends illustrated in Fig.~\ref{fig:drop_analysis_macro}. This analysis provides a direct estimate of localisation reliability loss when detection systems trained on clear daytime scenes are exposed to previously unseen adverse-weather environments.

For YOLOv8n, localisation remained stable under fog environments with no measurable degradation, while rain and snow introduced moderate reductions of 29.38\% and 23.53\%, respectively. The largest reduction appeared under night conditions, where performance decreased by 57.51\%, indicating strong sensitivity to illumination variation.

For YOLOv8m, performance remained nearly unchanged under fog conditions, with only 0.15\% reduction. Rain and snow produced comparable decreases of 24.57\% and 23.32\%, whereas night environments again resulted in the largest reduction, reaching 52.76\%.

For YOLO11m, localisation stability was preserved under fog conditions, followed by moderate reductions of 24.11\% under rain and 24.36\% under snow. Under night-time environments, performance declined more substantially to 53.58\%, reinforcing the dominant influence of low illumination on cross-weather reliability.

For Faster R-CNN, degradation remained limited under fog conditions and comparatively smaller under rain and snow environments, with reductions of 19.15\% and 16.05\%, respectively. However, night environments still produced the largest decrease, reaching 50.39\%.

For DETR, performance remained relatively stable under fog conditions but decreased more noticeably under rain environments to 35.39\%, followed by 24.40\% under snow. The strongest reduction occurred under night environments, where localisation reliability declined by 62.90\%.

Taken together, the degradation patterns shown in Table~\ref{tab:map50_drop_results} and Fig.~\ref{fig:drop_analysis_macro} indicate that night conditions consistently produced the largest performance reductions, followed by rain and snow environments, whereas fog conditions preserved localisation stability relative to the clear-weather baseline. This structured behaviour confirms that illumination variation represents the primary limiting factor affecting cross-weather perception reliability in camera-based autonomous driving systems.

\begin{table}[ht]
\centering
\caption{mAP50 degradation (\%) under adverse weather conditions.}
\label{tab:map50_drop_results}

\setlength{\tabcolsep}{2pt} 
\footnotesize 

\begin{tabular}{lcccc}
\toprule
Model & Fog Drop (\%) & Rain Drop (\%) & Snow Drop (\%) & Night Drop (\%) \\
\midrule

YOLOv8n      & 0.00 & 29.38 & 23.53 & 57.51 \\
YOLOv8m      & 0.15 & 24.57 & 23.32 & 52.76 \\
YOLO11m      & 0.00 & 24.11 & 24.36 & 53.58 \\
Faster R-CNN & 0.00 & 19.15 & 16.05 & 50.39 \\
DETR         & 2.97 & 35.39 & 24.40 & 62.90 \\

\bottomrule
\end{tabular}
\end{table}

\subsection{Object-category sensitivity under adverse weather conditions}

To investigate how environmental domain shift affects localisation reliability across semantic traffic participants, object-wise degradation relative to the clear-weather baseline is reported in Table~\ref{tab:classwise_map50_degradation_columns}. Unlike architecture-level Drop (\%) analysis, this evaluation reveals which object categories remain stable and which become vulnerable when detection systems trained under clear daytime conditions are deployed in unseen adverse-weather environments.

\begin{table*}[t]
\centering
\caption{Class-wise mAP50 degradation (\%) under adverse weather conditions.}
\label{tab:classwise_map50_degradation_columns}

\begin{tabular}{llcccc}
\toprule
Model & Object/Class & Fog Drop (\%) & Rain Drop (\%) & Snow Drop (\%) & Night Drop (\%) \\
\midrule

\multirow{4}{*}{YOLOv8n}
& Car    & 0.22 & 7.07 & 0.00 & 27.58 \\
& Bus    & 0.00 & 62.73 & 56.70 & 87.86 \\
& Truck  & 4.75 & 30.56 & 40.19 & 84.84 \\
& Person & 12.99 & 27.07 & 8.95 & 41.54 \\

\midrule

\multirow{4}{*}{YOLOv8m}
& Car    & 0.77 & 6.16 & 0.00 & 22.43 \\
& Bus    & 0.00 & 52.36 & 55.31 & 76.63 \\
& Truck  & 0.00 & 22.58 & 35.98 & 87.27 \\
& Person & 10.41 & 23.97 & 10.82 & 35.20 \\

\midrule

\multirow{4}{*}{YOLO11m}
& Car    & 0.64 & 5.76 & 0.00 & 21.61 \\
& Bus    & 0.00 & 51.99 & 58.24 & 77.65 \\
& Truck  & 0.00 & 21.33 & 38.13 & 90.96 \\
& Person & 10.22 & 23.58 & 9.95 & 34.58 \\

\midrule

\multirow{4}{*}{Faster R-CNN}
& Car    & 0.00 & 7.58 & 0.00 & 25.20 \\
& Bus    & 0.00 & 51.17 & 44.98 & 79.99 \\
& Truck  & 0.00 & 1.88 & 23.72 & 72.65 \\
& Person & 9.32 & 20.67 & 3.52 & 36.92 \\

\midrule

\multirow{4}{*}{DETR}
& Car    & 11.00 & 14.54 & 0.49 & 34.44 \\
& Bus    & 0.00 & 68.46 & 65.64 & 86.43 \\
& Truck  & 11.54 & 39.15 & 34.85 & 89.63 \\
& Person & 11.39 & 27.27 & 1.07 & 49.13 \\

\bottomrule
\end{tabular}
\end{table*}

Across all evaluated environments, the car category consistently exhibited the strongest localisation stability. Degradation remained negligible under fog and snow conditions and generally below 8\% under rain environments, while night-time conditions produced moderate reductions between approximately 21\% and 34\%. This behaviour indicates that large and structurally distinctive vehicles preserve sufficient geometric and appearance cues despite environmental variation.

In contrast, bus instances showed the largest sensitivity to adverse-weather deployment conditions. Although degradation remained minimal under fog environments, reductions exceeded 50\% under rain and snow conditions and increased beyond 75\% under night environments, indicating strong dependence on illumination quality and scene contrast for reliable localisation.

Truck detection demonstrated intermediate robustness behaviour. Performance reductions remained moderate under rain environments but increased substantially under snow and night conditions, with several cases exceeding 80\% degradation, suggesting increased sensitivity of medium-scale vehicle structures to reduced illumination and visibility.

Pedestrian localisation remained comparatively stable under fog and snow environments, where degradation typically remained below 15\%. However, reductions increased under rain environments and became more pronounced under night conditions, frequently exceeding 35\%, reflecting the combined influence of smaller object scale and reduced scene visibility.

Overall, the degradation patterns reported in Table~\ref{tab:classwise_map50_degradation_columns} demonstrate that cross-weather localisation reliability is strongly object dependent rather than uniformly affected across traffic participants. Large structured vehicles such as cars remain comparatively stable across environmental conditions, whereas buses and trucks exhibit substantial sensitivity to illumination variation and contrast reduction, and pedestrians show intermediate robustness with degradation increasing primarily under night-time deployment scenarios. 
These results indicate that object geometry, scale, and illumination contrast jointly determine localisation stability under cross-weather domain shift, highlighting the importance of object-specific robustness analysis for reliable autonomous driving perception in real-world environmental conditions.
\subsection{Architecture-level Cross-weather Robustness Analysis Using the Weather Robustness Index (WRI)}

While Drop (\%) analysis quantifies performance reduction under individual adverse-weather environments, it does not indicate how much localisation capability is preserved overall relative to the clear-weather baseline. To address this limitation, cross-weather generalisation was further evaluated using the Weather Robustness Index (WRI), defined in Eq.~(\ref{eq:wri_metric}) as the ratio between average adverse-weather performance and clear-weather performance. This normalised measure provides a direct estimate of robustness retention after environmental domain shift.

\begin{figure}[ht]
\centering
\includegraphics[width=\columnwidth]{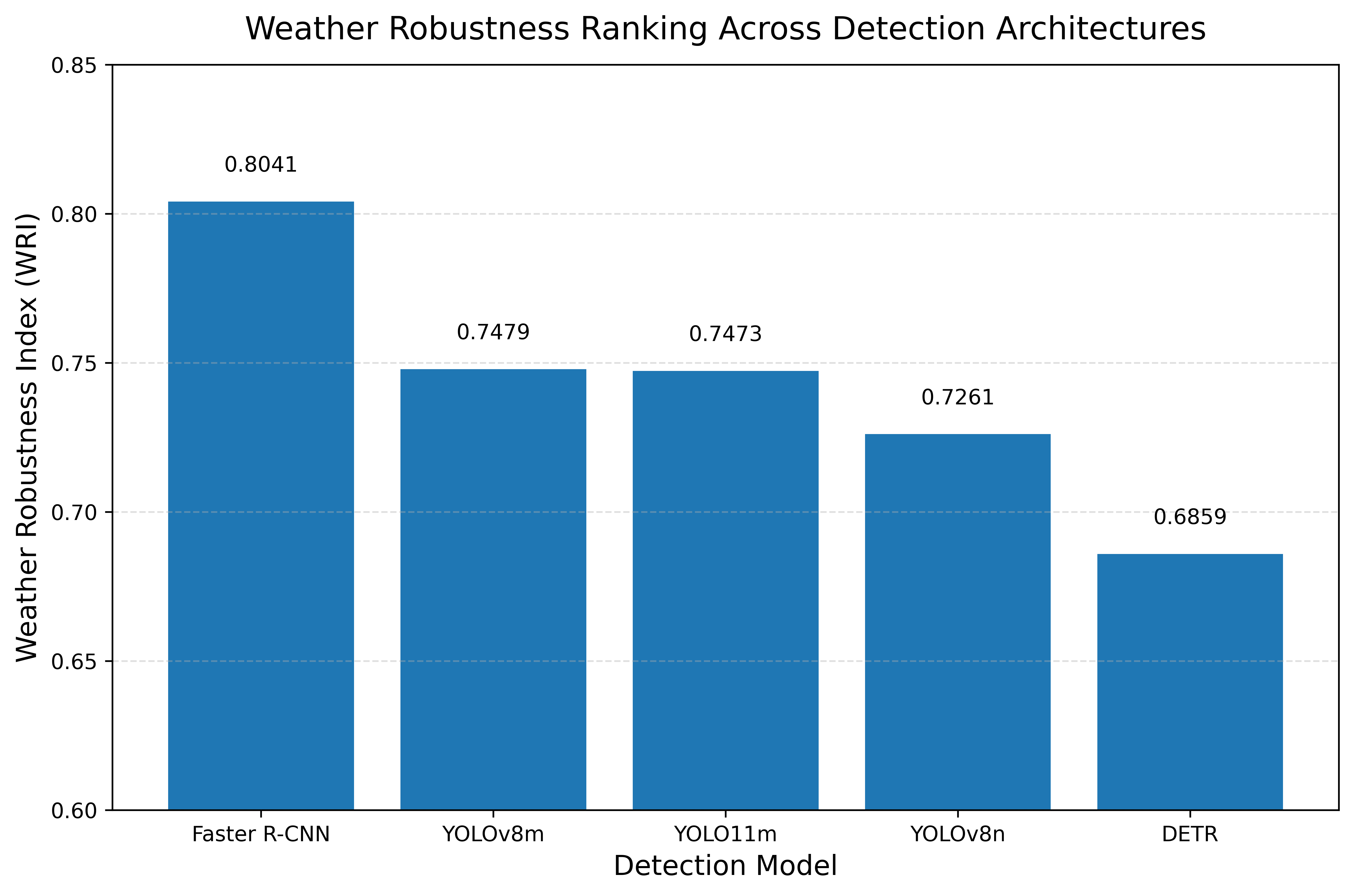}
\caption{Weather Robustness Index (WRI) ranking across detection architectures.}
\label{fig:wri_ranking}
\end{figure}

The resulting WRI values are reported in Table~\ref{tab:wri_results}, with the robustness ranking illustrated in Fig.~\ref{fig:wri_ranking}. Unlike Drop (\%), which reflects condition-specific degradation, WRI summarises the overall stability of a detection pipeline across multiple unseen environments. Higher WRI values therefore indicate stronger preservation of localisation reliability under cross-weather deployment conditions.

\begin{table*}[t]
\centering
\caption{Weather Robustness Index (WRI) across detection architectures.}
\label{tab:wri_results}

\begin{tabular}{lcccc}
\toprule
Model & Clear mAP50 & Avg ACDC mAP50 & WRI & Interpretation \\
\midrule

YOLOv8n      & 0.6022 & 0.4373 & 0.7261 & Moderately Weak \\
YOLOv8m      & 0.6698 & 0.5010 & 0.7479 & Moderately Weak \\
YOLO11m      & 0.6818 & 0.5094 & 0.7473 & Moderately Weak \\
Faster R-CNN & 0.5594 & 0.4498 & 0.8041 & Moderately Robust \\
DETR         & 0.2795 & 0.1917 & 0.6859 & Weak \\

\bottomrule
\end{tabular}
\end{table*}

For YOLOv8n, the WRI value of 0.7261 indicates that approximately three quarters of clear-weather localisation capability is retained under adverse-weather deployment conditions, confirming moderate stability across environmental shifts and remaining consistent with the larger reductions previously observed under night environments.

For YOLOv8m, the WRI value of 0.7479 indicates improved retention of localisation performance across unseen weather conditions, suggesting stronger resistance to visibility variation compared with the lightweight configuration.

For YOLO11m, the comparable WRI value of 0.7473 indicates similar robustness behaviour under environmental variation. The close agreement between YOLOv8m and YOLO11m suggests that robustness across adverse-weather environments is influenced more strongly by scene visibility conditions than by architectural scaling alone.

For Faster R-CNN, the highest WRI value of 0.8041 indicates the strongest preservation of localisation performance relative to the clear-weather baseline, demonstrating improved stability across adverse-weather environments within the evaluated detection pipelines.

For DETR, the WRI value of 0.6859 indicates comparatively weaker preservation of localisation capability under cross-weather deployment conditions, reflecting increased sensitivity to visibility variation and illumination changes previously identified through Drop (\%) analysis.
The robustness ranking presented in Fig.~\ref{fig:wri_ranking} shows that cross-weather localisation reliability remains within the moderately weak to moderately robust range across the evaluated detection pipelines. These results confirm that environmental domain shift systematically reduces detection performance, with illumination variation representing the dominant limiting factor when models trained on clear-weather data are deployed under unseen adverse-weather conditions.

These robustness trends also highlight the practical trade-off between computational efficiency and deployment reliability in autonomous driving perception systems. Lightweight one-stage detectors provide favourable real-time efficiency but exhibit moderate robustness under cross-weather domain shift. In contrast, the two-stage detection architecture demonstrates stronger robustness stability despite higher computational complexity, whereas the transformer-based detector shows increased sensitivity to environmental variation under the evaluated deployment setting. These observations indicate that selecting perception architectures for real-world autonomous driving systems requires balancing inference efficiency with robustness stability rather than relying solely on clear-weather performance benchmarks.

\begin{table}[ht]
\centering
\caption{ECE across clear and cross-weather evaluation conditions (lower is better).}
\label{tab:ece_results}
\begin{tabular}{lccccc}
\hline
Model & Clear & Fog & Rain & Night & Snow \\
\hline
YOLOv8n & 0.0163 & 0.0125 & 0.0144 & 0.0140 & 0.0154 \\
YOLOv8m & 0.0204 & 0.0125 & 0.0164 & 0.0142 & 0.0174 \\
YOLO11m & 0.0214 & 0.0187 & 0.0162 & 0.0144 & 0.0161 \\
Faster R-CNN & 0.2017 & 0.2442 & 0.2843 & 0.2557 & 0.2367 \\
DETR & 0.4383 & 0.5314 & 0.5484 & 0.5790 & 0.5392 \\
\hline
\end{tabular}
\end{table}

\subsection{Prediction Confidence Reliability Analysis}

To assess prediction confidence reliability under environmental domain shift, calibration behaviour was evaluated using Expected Calibration Error (ECE), defined in
Eq.~(\ref{eq:ece_metric}). Unlike detection accuracy metrics such as mAP, which measure localisation and classification performance, ECE quantifies the agreement between predicted confidence scores and empirical detection correctness. This enables additional interpretation of whether detection models remain trustworthy when deployed under unseen adverse-weather conditions.

Table~\ref{tab:ece_results} presents calibration performance across clear-weather and cross-weather evaluation environments. Lower ECE values indicate stronger agreement between predicted confidence and actual detection correctness.

The findings reveal that the YOLO-based detectors have a consistently low calibration error in both clear-weather and adverse-weather conditions, and the ECE values do not exceed 0.03 in all the conditions of the evaluation. This behaviour implies steady consensus between prediction confidence and empirical detection correctness even when visibility is decreased and there is a change in illumination.

Conversely, Faster R-CNN has significantly higher calibration error in all settings where ECE is 0.2017 in clear-weather conditions and 0.2843 in rain conditions and is higher in all the other cases, in fog, snow, and night conditions. This implies a low confidence reliability in cases where the operating environment is not similar to the training domain.

The transformer-based DETR detector demonstrates the largest calibration mismatch among the evaluated architectures. ECE increases from 0.4383 in clear-weather evaluation to 0.5790 under night-time conditions and remains consistently above 0.53 across all adverse-weather environments. These results suggest that prediction confidence becomes progressively less reliable under environmental domain shift, particularly in low-visibility scenarios.

Overall, the calibration analysis complements detection accuracy and robustness results by showing that lightweight one-stage detectors maintain more consistent confidence reliability under cross-weather deployment conditions, whereas two-stage and transformer-based detectors exhibit larger confidence discrepancies when evaluated outside the training domain.

\subsection{Qualitative Detection Behaviour Across Weather Conditions}

\begin{figure*}[ht]
\centering
\includegraphics[width=\textwidth]{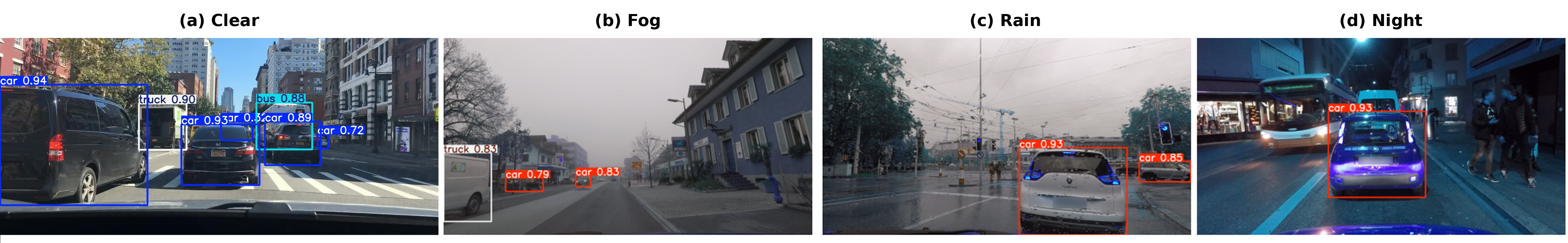}
\caption{Qualitative YOLOv8n detection performance across weather conditions, highlighting stronger degradation at night.}
\label{fig:qualitative_weather}
\end{figure*}

To complement the quantitative evaluation, representative qualitative detection examples across clear, fog, rain, and night driving environments are presented in Fig.~\ref{fig:qualitative_weather}. These examples provide visual insight into cross-weather domain generalisation behaviour and support interpretation of the degradation trends observed in the performance tables.

Under clear daytime conditions (Fig.~\ref{fig:qualitative_weather}(a)), the detector accurately localises multiple vehicle categories, including cars, buses, and trucks, with stable bounding box placement and consistent prediction confidence, confirming reliable in-domain detection performance.

In foggy conditions (Fig.~\ref{fig:qualitative_weather}(b)), the detector continues to identify major vehicle classes despite reduced visibility. Although fewer objects are detected compared to clear conditions, large and structurally distinct vehicles remain detectable, indicating partial robustness to moderate visibility degradation.

Under rain conditions (Fig.~\ref{fig:qualitative_weather}(c)), detection becomes more selective. Nearby vehicles remain identifiable, but the number of detected instances decreases relative to clear and fog environments, consistent with the quantitative performance reduction reported earlier.

Night-time conditions (Fig.~\ref{fig:qualitative_weather}(d)) present the most challenging scenario. While a nearby vehicle is detected, several surrounding objects such as pedestrians and buses remain missed, illustrating the strong influence of illumination variation on detection reliability and prediction confidence stability.

Taken together, these qualitative observations confirm that cross-weather performance degradation is driven by both visibility reduction and illumination variation, with night-time conditions producing the most pronounced instability. This behaviour is consistent with the Drop (\%) analysis, Weather Robustness Index results, and confidence reliability trends observed in the calibration evaluation.
\section{Conclusion and Future Work}
\label{sec:conclusion}
This study presented a controlled source-only evaluation of object detection robustness under environmental domain shift for autonomous driving perception systems. Detection architectures trained exclusively on clear daytime driving scenes were evaluated across unseen fog, rain, snow, and night environments to assess deployment-oriented reliability under realistic cross-weather transitions. Experimental results showed consistent performance degradation across all adverse-weather conditions, with the largest reduction (62.90\%) observed under night-time environments, confirming illumination variation as the primary limiting factor affecting cross-weather localisation stability.
Class-level analysis demonstrated that robustness is strongly object dependent, with car instances remaining comparatively stable, whereas bus and truck categories showed substantial sensitivity to reduced visibility and illumination changes. Architecture-level comparison indicated that Faster R-CNN preserved the highest proportion of localisation capability across adverse-weather environments (WRI = 0.8041), while DETR exhibited the lowest robustness retention (WRI = 0.6859). Confidence calibration analysis further showed that YOLO-based detectors maintained consistently low calibration error across evaluation environments, indicating more reliable prediction confidence under cross-weather deployment conditions.

These findings highlight that detection accuracy measured under favourable conditions alone does not adequately reflect real-world deployment reliability. The proposed source-only cross-weather evaluation framework enables structured analysis of architecture-level robustness, object-category sensitivity, and prediction confidence reliability under heterogeneous weather conditions without target-domain supervision.

Future work will extend this evaluation to additional adverse-weather datasets and investigate domain generalisation strategies for improving perception robustness under safety-critical autonomous driving deployment conditions.

\bibliographystyle{IEEEtran}
\bibliography{bibliography}




\section{Biography Section}

\vspace{-1.33cm}
\begin{IEEEbiography}[{\includegraphics[width=1in,height=1.25in,clip,keepaspectratio]{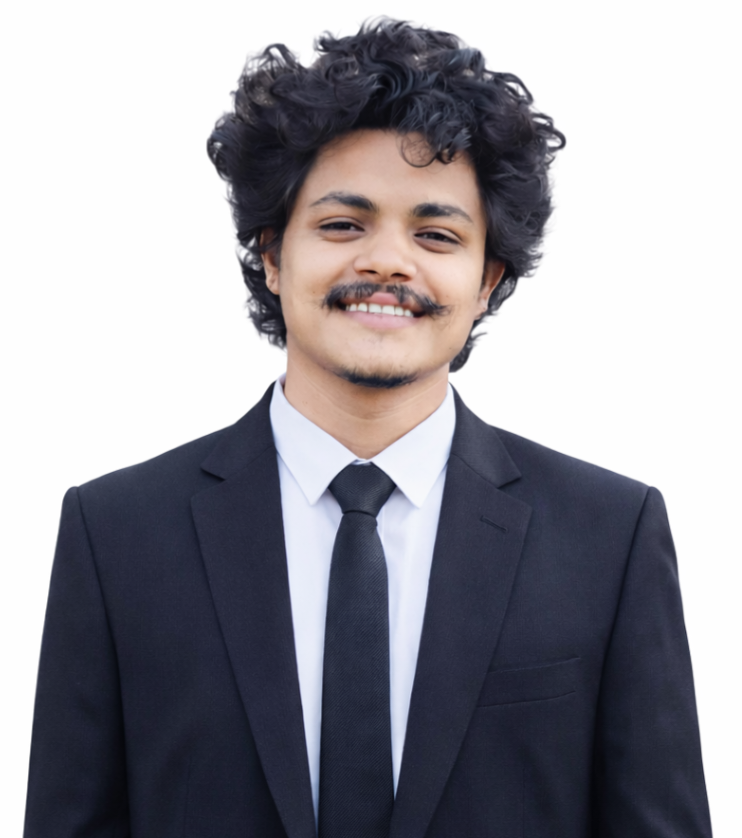}}]{Munshaibur Rahman Mahin}
 is currently pursuing the B.Sc. degree in Computer Science and Engineering at Sylhet Engineering College, Shahjalal University of Science and Technology (SUST), Bangladesh. His research interests include computer vision, deep learning for medical image analysis, natural language processing, and machine learning. He received the National Award in 2018 from the Prime Minister of Bangladesh in recognition of his academic achievements. His current research focuses on developing efficient and reliable intelligent learning systems for practical applications in vision-based and language-based artificial intelligence.
\end{IEEEbiography}

\vspace{-25pt}

\begin{IEEEbiography}[{\includegraphics[width=1in,height=1.25in,clip,keepaspectratio]{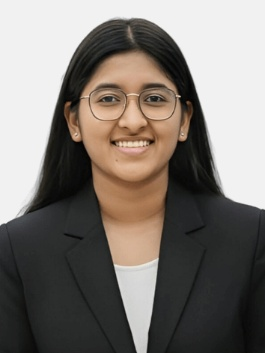}}]{Isha Das}
 is an Electrical Engineering student at the University of Toronto focused on embedded systems and power electronics. She is interested in designing efficient hardware systems and applying engineering principles to real-world energy and electronics challenges.
\end{IEEEbiography}

\vspace{-25pt}

\begin{IEEEbiography}[{\includegraphics[width=1in,height=1.25in,clip,keepaspectratio]{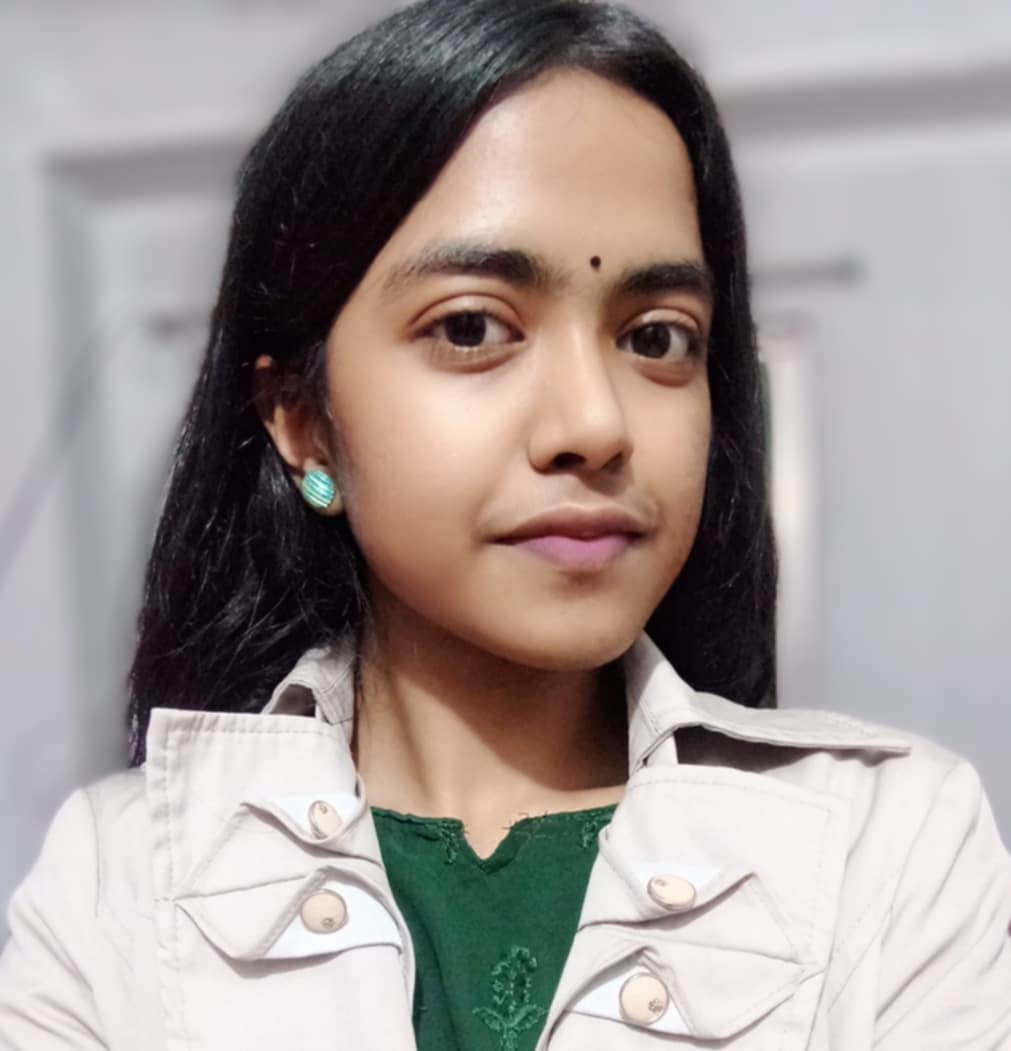}}]{Most. Sonia Islam}
received the B.Sc. degree in Computer Science and Engineering from the Bangladesh University of Business and Technology (BUBT), Dhaka, Bangladesh, in April 2026.
Her research interests include deep learning, computer vision, and medical image analysis.
\end{IEEEbiography}

\vspace{-20pt}
\begin{IEEEbiography}[{\includegraphics[width=1in,height=1.25in,clip,keepaspectratio]{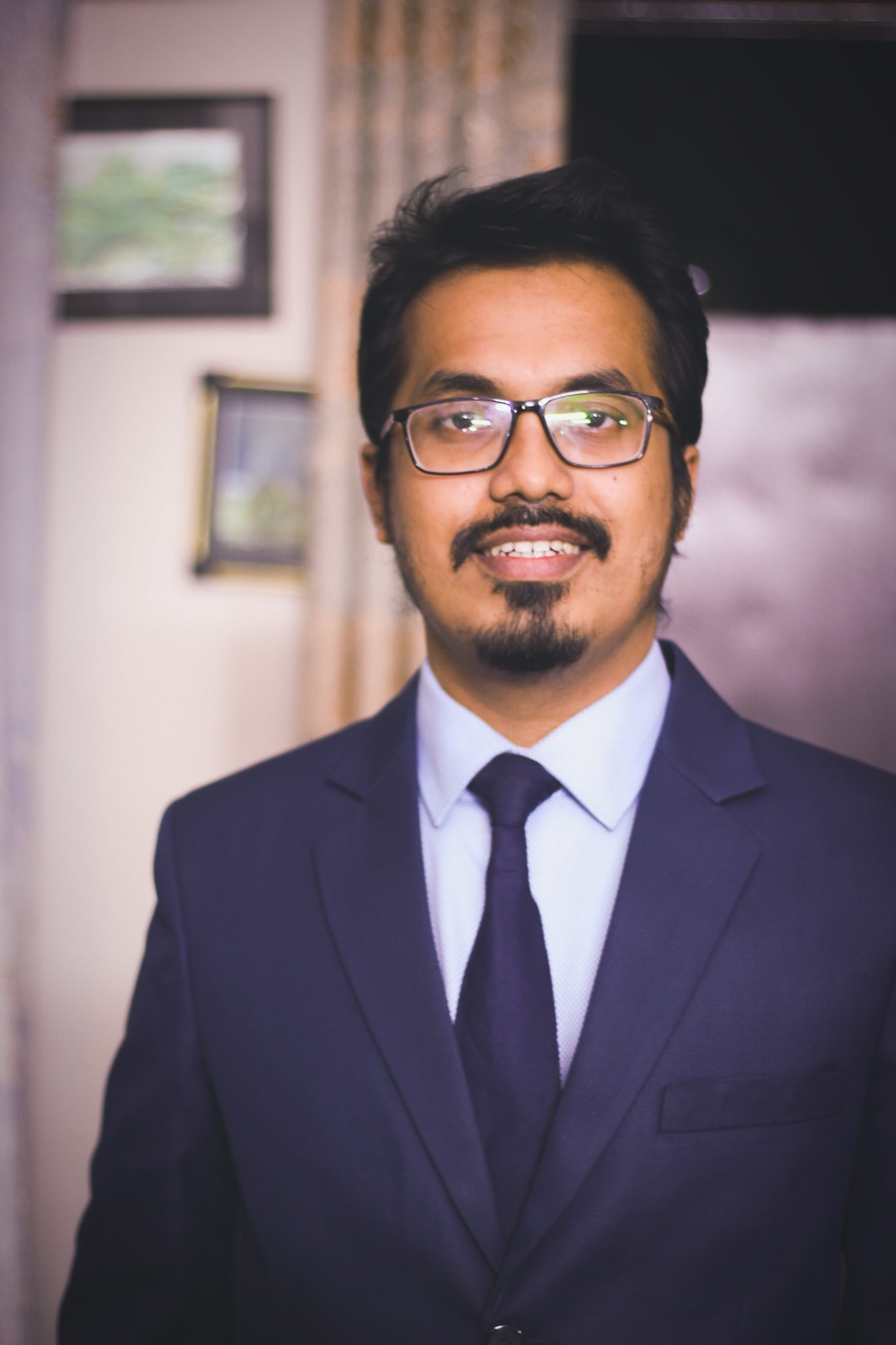}}]{Mohammad Sakib Mahmood} is a graduate student in the Department of Computer Science at Missouri State University, Springfield, MO, U.S.A and an active researcher in the field of Artificial Intelligence. His research interests broadly encompass machine learning, deep learning, computer vision, reinforcement learning, and system security. He is particularly interested in developing intelligent systems that can learn, adapt, and make decisions in complex environments.

His research interests include machine learning, deep learning, computer vision, reinforcement learning, and system security, with a particular focus on developing intelligent and adaptive systems for real-world applications. His recent work explores the integration of advanced AI techniques in areas such as computer vision and reinforcement learning-based environments. He also serves as a Program Committee Member (Reviewer) for the 38th IEEE International Conference on Tools with Artificial Intelligence (ICTAI 2026), held in Boca Raton, Florida, USA, where he reviewed research submissions in Artificial Intelligence and Machine Learning. In this role, he provided detailed technical evaluations, constructive feedback, and recommendations to support the peer-review process.

He is passionate about bridging theoretical research and practical applications, with the long-term goal of contributing to impactful advancements in intelligent systems.
\end{IEEEbiography}

\begin{IEEEbiography}[{\includegraphics[width=1in,height=1.25in,clip,keepaspectratio]{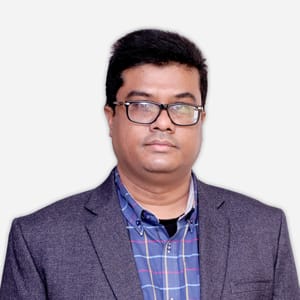}}]
{Dr. M. F. MRIDHA}~(Senior Member IEEE, Professional Member ACM) is currently working as a Professor in the Department of Computer Science, American International University-Bangladesh (AIUB). His research interests include Artificial Intelligence (AI), Computer Vision (CV), Machine Learning(ML), Deep Learning(DL), and Natural Language Processing (NLP) etc. He achieved the top 5 most productive researchers in Bangladesh by (Scopus/ Elsevier) and secured the Top 2\% researcher in the field of Computer Science and Engineering in 2024 and 2025.
\end{IEEEbiography}

\vspace{-1cm}
\begin{IEEEbiography}[{\includegraphics[width=1in,height=1.25in,clip,keepaspectratio]{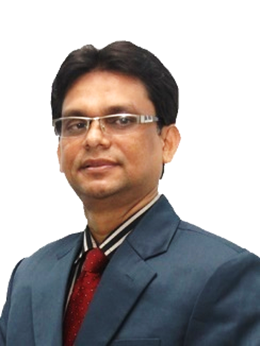}}]
{Dr. Md. Jakir Hossen (Senior Member, IEEE)}~is currently working as an Associate Professor in the Department of Robotics and Automation, Faculty of Engineering and Technology, Multimedia University, Melaka, Malaysia. He received the master’s degree in communication and network engineering from Universiti Putra Malaysia, Malaysia in 2003. He received the PhD degree in smart technology and robotic engineering from Universiti Putra Malaysia, Malaysia in 2012. His research interests are the applications of artificial intelligence techniques in data analytics, robotics control, data classifications and predictions.
\end{IEEEbiography}

\vfill

\end{document}